\documentclass[10pt,twocolumn,letterpaper]{article}

\usepackage[pagenumbers]{cvpr} 

\usepackage{graphicx}
\usepackage{amsmath}
\usepackage{amssymb}
\usepackage{booktabs}
\usepackage{gensymb}
\usepackage{multirow}
\usepackage{rotating}
\usepackage{fancyhdr}
\usepackage{latexsym}
\usepackage{url}
\usepackage{algorithm}
\usepackage{algorithmic}
\usepackage{bm}
\usepackage{stmaryrd}
\usepackage{dsfont}
\usepackage{epstopdf}
\usepackage{booktabs}
\usepackage{dashrule}
\usepackage{soul}
\usepackage{color}
\usepackage[table,x11names]{xcolor}

\usepackage{ragged2e}
\usepackage[pagebackref,breaklinks,colorlinks]{hyperref}

\usepackage[accsupp]{axessibility}

\usepackage[capitalize]{cleveref}
\crefname{section}{Sec.}{Secs.}
\Crefname{section}{Section}{Sections}
\Crefname{table}{Table}{Tables}
\crefname{table}{Tab.}{Tabs.}

\newcommand{\blue}[1]{\textcolor{blue}{\bf{#1}}}
\newcommand{\red}[1]{\textcolor{red}{\bf{#1}}}

\newcommand{\ours}{NOC-REK}

\begin{document}


\title{NOC-REK: Novel Object Captioning with Retrieved Vocabulary \\ from External Knowledge}

\author{Duc Minh Vo \\
The University of Tokyo, Japan\\
{\tt\small vmduc@nlab.ci.i.u-tokyo.ac.jp}
\and
Hong Chen\\
The University of Tokyo, Japan\\
{\tt\small chen@nlab.ci.i.u-tokyo.ac.jp}
\and
Akihiro Sugimoto\\
National Institute of Informatics, Japan\\
{\tt\small sugimoto@nii.ac.jp}
\and
Hideki Nakayama\\
The University of Tokyo, Japan\\
{\tt\small nakayama@ci.i.u-tokyo.ac.jp}
}

\maketitle

\begin{abstract}

Novel object captioning aims at describing objects absent from training data, with the key ingredient being the provision of object vocabulary to the model.
Although existing methods heavily rely on an object detection model, we view the detection step as vocabulary retrieval from an external knowledge in the form of embeddings for any object's definition from Wiktionary, where we use in the retrieval image region features learned from a transformers model.
We propose an end-to-end \textbf{N}ovel \textbf{O}bject \textbf{C}aptioning with \textbf{R}etrieved vocabulary from \textbf{E}xternal \textbf{K}nowledge method (\ours{}), which simultaneously learns vocabulary retrieval and caption generation, successfully describing novel objects outside of the training dataset.
Furthermore, our model eliminates the requirement for model retraining by simply updating the external knowledge whenever a novel object appears. 
Our comprehensive experiments on held-out COCO and Nocaps datasets show that our \ours{} is considerably effective against SOTAs.

\end{abstract}

\section{Introduction}

\begin{figure}[t]
	\centering
	\includegraphics[width=\linewidth]{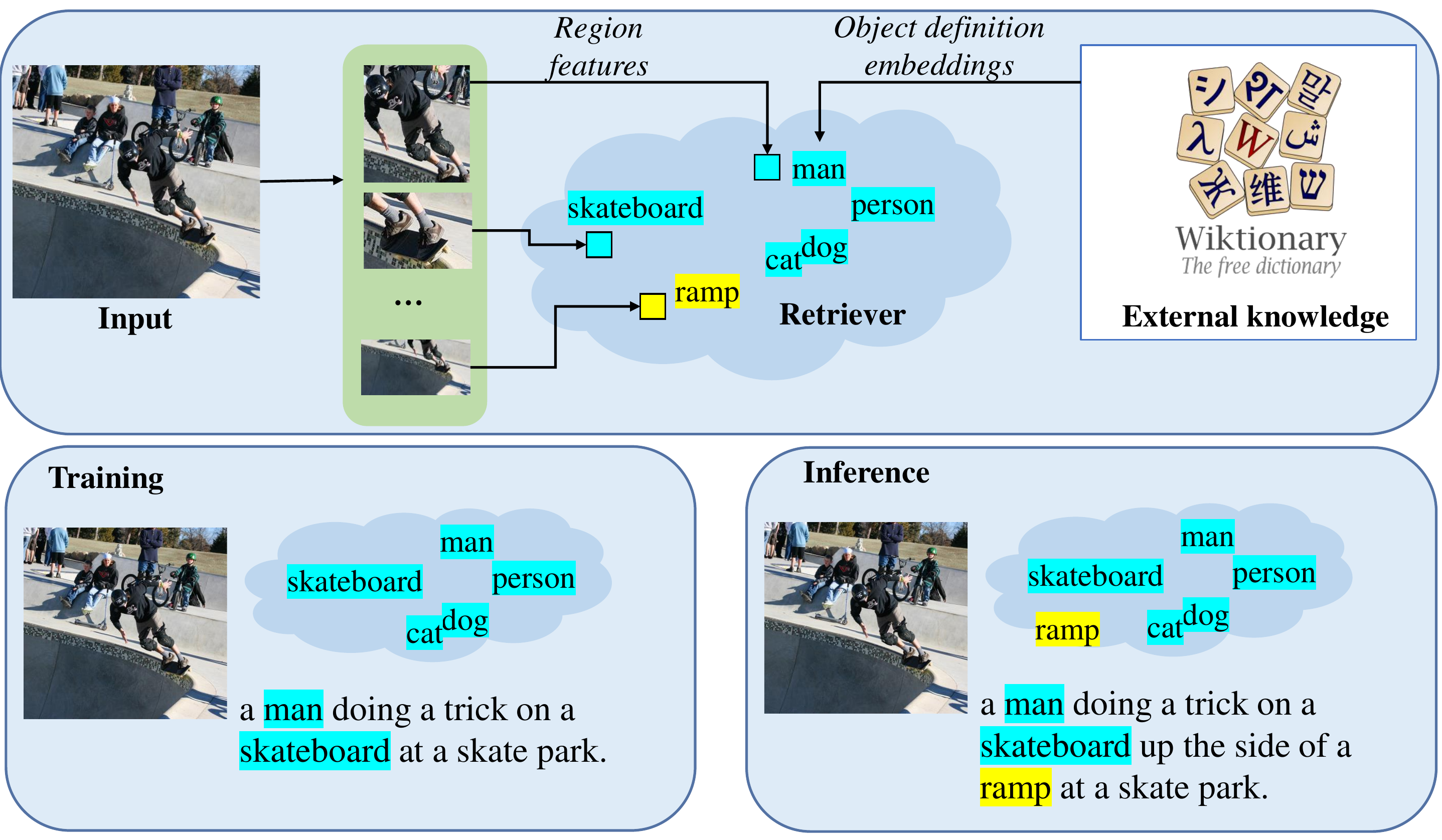}
	\caption{(upper part) Our \ours{} first retrieves (object) vocabulary by computing similarity of region features and object definition embeddings from an external knowledge. The caption is then generated using those retrieved vocabulary and region features. 
	(below part) At training phase, the model is trained on  paired image-caption data and external knowledge where both of them cover a limited number of objects (blue rectangles and texts). 
	At inference time, our model generally describes novel objects by simply updating the external knowledge (yellow text) without any extra training.}  
	\label{fig:overview}
\end{figure}

Describing \textit{novel objects} not observable in training data expands the real-world applications of image captioning models.
As we progress toward the new goal, the image captioning task shifts to novel object captioning.
The conventional image captioning models~\cite{Anderson2017up-down,vinyals2014neural,Lu2018nbt}, however, fail to describe novel objects because they only learn the correspondence between images and sentences provided in the image--caption pair data.
To overcome this limitation, the most straightforward way is to collect more data~\cite{hendricks16dcc,venugopalan17noc} so that images and captions contain the novel objects that we need; and then retrain the whole system.
The newly collected data will provide the model with at least object names that appear in the given image~\cite{Mikihiro2020ove}.
The caption will then include novel objects.
Even if such methods~\cite{hendricks16dcc,venugopalan17noc} are effective to some extent, the data collection process is highly expensive and time-consuming.
How to efficiently provide object vocabulary to the model is, therefore, desired.

A less expensive yet more effective approach~\cite{wu2018dnoc,Berkan2019zsc,Mikihiro2020ove,xianyu2021anoc,li2020oscar,hu2021vivo,zhang2021vinvl} that relies on object detection model has been proposed, demonstrating breakthroughs in this task.
Despite the fact that modern object detection models (e.g., Faster RCNN~\cite{ren2015faster}, zero-shot object detection~\cite{Demirel2018ZeroShotOD}) can recognize a wide range of objects including novel ones, using  object detection in novel object captioning models~\cite{hendricks16dcc,venugopalan17noc,Lu2018nbt,wu2018dnoc,Berkan2019zsc,Mikihiro2020ove,xianyu2021anoc,li2020oscar,hu2021vivo,zhang2021vinvl} brings a new challenge.
Additional efforts are still required to improve a portion of the detection model in order to broaden the model's knowledge of novel objects.
Such efforts become massive as the number of objects is unlimited in the wild.
This study aims to reduce such workload whenever a new object becomes available by providing object vocabulary during inference without additional training or data.

Humans usually build our knowledge of objects by defining any object at a concrete level based on its appearance (e.g., \textit{cat}, \textit{dog}) or an abstract level in conjunction with other objects (e.g., \textit{kitchen}).
Therefore, we can easily match an object and its definition, answering its name, regardless of whether we have previously encountered the object.
Motivated by that fact, we simplify the object detection by viewing it as vocabulary retrieval from a set of objects' definitions (hereafter referred to as external knowledge).
As shown in Fig.~\ref{fig:overview}, we match region features and object definition embeddings using their similarity, returning relevant vocabulary.
Note that the dataset is used to train our model only to cover a limited number of objects.
When a new object appears after training, we simply update the external knowledge at a much lower cost than previous methods.
Our model, therefore, can generally describe any image that contains the novel object (e.g., \textit{a man doing a trick on a skateboard up the side of a \textbf{ramp} at a skate park} in Fig.~\ref{fig:overview}).

We propose \textbf{N}ovel \textbf{O}bject \textbf{C}aptioning with \textbf{R}etrieved vocabulary from \textbf{E}xternal \textbf{K}nowledge method, abbreviated as \ours{}.
Inspired by significant advances of transformers~\cite{Vaswani2017attention} in object detection~\cite{nicolas2020detr} and vision-language models~\cite{li2020oscar,hu2021vivo,zhang2021vinvl}, \ours{} makes full use of a shared-parameter transformers model to unify vocabulary retrieval and caption generation steps in an end-to-end manner.
More specifically, we prepare the external knowledge by using object definitions from Wiktionary\footnote{\url{https://en.wiktionary.org/wiki/Wiktionary:Main_Page}} and a pre-trained BERT model~\cite{devlin-etal-2019-bert}.
\ours{} first learns region features from a set of Faster R-CNN~\cite{ren2015faster} regions of interest (ROIs).
Then, we perform vocabulary retrieval using object definition embeddings (in the external knowledge) and region features by computing their similarity.
Finally, the caption is generated using the retrieved vocabulary and region features (Fig.~\ref{fig:overview}).
We train the vocabulary retrieval using Hungarian loss~\cite{nicolas2020detr} with our modification, while at the first training stage, we use cross-entropy, and at the second training stage, we use SCST~\cite{Rennie2017SCST} with a reward for the appearance of novel objects to train caption generation.
Our contributions are:
\begin{itemize}
    \item We simplify the object detection step used in the image captioning model by viewing it as vocabulary retrieval from external knowledge.
    \item We propose an end-to-end \ours{} model which retrieves vocabulary from external knowledge and generates captions using shared-parameter transformers. 
    \item Our method provides object vocabulary during inference, effectively eliminating the necessity for either retraining model or extra image and/or caption data when a novel object appears.
\end{itemize}

\section{Related work}

\subsection{Novel object captioning} 

This task aims at describing objects unseen during the training phase (called novel objects) where many methods~\cite{hendricks16dcc,venugopalan17noc,Lu2018nbt,wu2018dnoc,Berkan2019zsc,Mikihiro2020ove,xianyu2021anoc,li2020oscar,hu2021vivo,zhang2021vinvl} have been proposed.
Hendricks et al. (DCC)~\cite{hendricks16dcc} and Venugopalan et al. (NOC)~\cite{venugopalan17noc} purposely use unpaired labeled image and sentence data to learn semantically visual concepts.
Lu et al. (NBT)~\cite{Lu2018nbt}, Wu et al. (DNOC)~\cite{wu2018dnoc}, and Demirel et al. (ZSC)~\cite{Berkan2019zsc} fill the generated template sentence with objects detected by object/novel object detectors.
Tanaka et al. (OVE)~\cite{Mikihiro2020ove} propose a low-cost method to expand word embeddings from a few images of the novel objects.
Chen et al. (ANOC)~\cite{xianyu2021anoc} combine object detector and human attention to identify novel objects.
On the other hand, Li et al. (Oscar)~\cite{li2020oscar}, Hu et al. (VIVO)~\cite{hu2021vivo} and Zhang et al. (VinVL)~\cite{zhang2021vinvl} pre-train large-scale vision-language transformers models and then finetune the pre-trained model to adopt downstream tasks.

Generally, the above mentioned methods follow a two-step approach where the detection step is prior to the caption generation step.
The former step leverages off-the-shelf object detectors such as Faster RCNN~\cite{ren2015faster}, zero-shot object detection~\cite{Demirel2018ZeroShotOD} to identify objects, which requires additional training for new objects.
The latter step employs either LSTM~\cite{hendricks16dcc,venugopalan17noc,Lu2018nbt,wu2018dnoc,Berkan2019zsc}
or transformers~\cite{li2020oscar,hu2021vivo,zhang2021vinvl} in which transformers generate better captions.
The two steps are trained independently, requiring much efforts to include the novel objects into the caption~\cite{venugopalan17noc}.

Different from the aforementioned methods, we view the detection step as vocabulary retrieval, allowing our method to be trained in an end-to-end way on a transformers model.
Moreover, our method does not require any re-training or additional data even when a novel object arises.

\subsection{Transformers-based vision-language pre-training models}

Recent vision-language methods~\cite{li2020oscar,hu2021vivo,zhang2021vinvl} are BERT-like~\cite{devlin-etal-2019-bert} models which successfully learn the vision-language cross-modal by using a concatenated-sequence of words---object tags---visual regions as its input,
showing breakthroughs in many vision-language tasks such as image captioning, image-text retrieval.
In fact, those models are pre-trained on a large image-text corpus that probably contains novel objects, so their performance on the novel object captioning task is doubly ambiguous.
Like~\cite{li2020oscar,hu2021vivo,zhang2021vinvl}, our method also uses a concatenated-sequence.
However, we do not pre-train the model to avoid biases.

\section{Proposed \ours}

\subsection{Idea of \ours{}}

We simplify the object detection step, as discussed above, by viewing it as vocabulary retrieval from external knowledge.
We then unify vocabulary retrieval and caption generation in an end-to-end manner. 
Two critical challenges arise here: (1) how to realize a knowledge-based vocabulary retrieval and (2) how to design an architecture that allows the entire model to be jointly end-to-end trained.

In order to tackle challenge (1), the straightforward solution is to compute the similarity of image features and object definition embeddings from external knowledge.
The main difficulty of training is to score retrieved vocabulary with respect to ground-truth objects.
This is because the order of the ground-truth objects and that of the retrieved vocabulary differs.
We thus use the Hungarian loss proposed in~\cite{nicolas2020detr} to train our model.
However, unlike~\cite{nicolas2020detr}, which rejects objects that are not in training data, we introduce a simple yet effective modification to ensure that the model adopts objects that are not in ground-truth.

\begin{figure}[t]
	\centering
	\includegraphics[width=\linewidth]{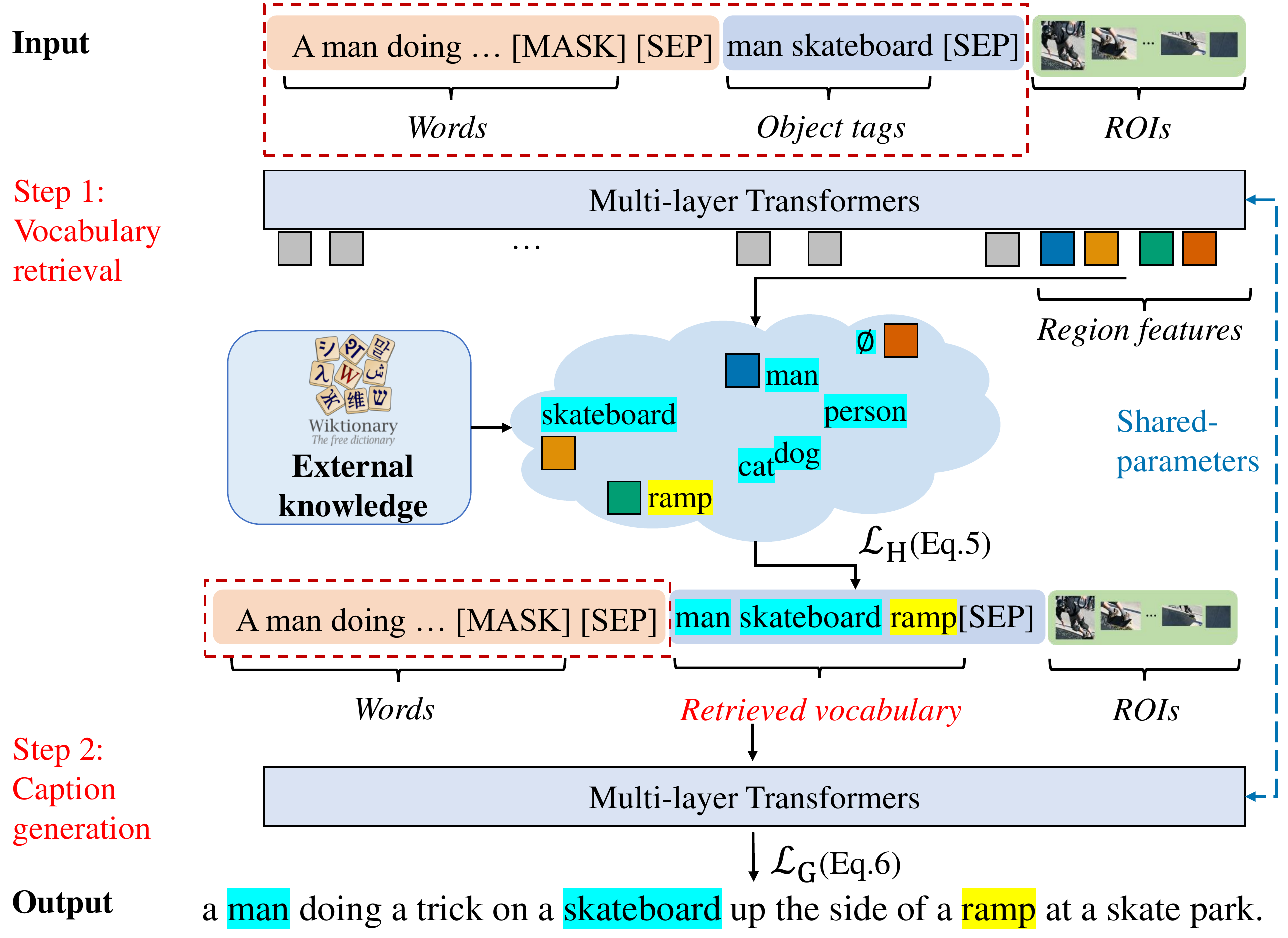}
	\caption{The overall pipeline of our proposed \ours{}, which includes two steps: Vocabulary retrieval and Caption generation. Our model is made up of multi-layer transformers with shared parameters. The transformers is first used in the knowledge-based vocabulary retrieval step, where it return the objects in the given image. For the caption generation step, we use the same transformers as before. After training, the dashed red parts will be removed. $\mathtt{[MASK]}$ and $\mathtt{[SEP]}$ are special tokens to mask some tokens in the sequence of words and separate the functions of different inputs.} 
	\label{fig:framework}
\end{figure}

Transformers-based model shows impressive results in object detection~\cite{nicolas2020detr}, motivating us to build our retrieval upon transformers.
At the same time, following the success of transformers-based vision-language models~\cite{li2020oscar,hu2021vivo,zhang2021vinvl} in image captioning, we adopt the model proposed in~\cite{zhang2021vinvl} for our caption generation where the input is a concatenated-sequence of words---object tags---ROIs.
Therefore, we propose using shared-parameter transformers for both vocabulary retrieval and caption generation, thereby answering (2) (Fig.~\ref{fig:framework}).
The usage of shared-parameters reduces training costs while improving the learning of the vocabulary retrieval step.
This is thanks to the ability of image features to cross-attend to language information, resulting in better alignment between vision and language spaces prior to performing the retrieval step.
In what follows, we describe the precise details of our method as the preceding discussions.

\subsection{Knowledge-based vocabulary retrieval} \label{sec:retrieval}

We conceptualize and mathematically formulate our knowledge-based vocabulary retrieval in this section.
The technical integration of vocabulary retrieval and caption generation is presented in Section~\ref{sec:architecture}.

\noindent
\textbf{Selection of external knowledge.}
We use a free dictionary Wiktionary that describes all words using definitions and descriptions to build our external knowledge.
We first crawl the definition of each object name in the target dataset. 
Assuming that we obtain $M$ vocabulary (i.e., object names and our defined 'no object') with their corresponding definitions, we employ a pre-trained BERT~\cite{devlin-etal-2019-bert} to embed the definition of each word $v$ into a embedding with the size of $1 \times 768$: $\mathbf{d} = \operatorname{BERT}(v)$ ('no object' is set to the vector of all-zeroes.)
As a result, each word will be presented as a key-value pair $(\mathbf{d}, v)$.
The external knowledge now consists of $M$ key-value pairs.

\noindent
\textbf{Object detection as vocabulary retrieval.}
We first define the similarity between the region feature and definition embedding $\mathbf{d}$ in the external knowledge:
\begin{equation}
    \operatorname{sim}(\mathbf{r}, \mathbf{d}) = \frac{\mathbf{r}^{\top} \mathbf{d}}{\left \| \mathbf{r} \right \| \left \| \mathbf{d} \right \|},
    \label{eq:similarity}
\end{equation}
%
where $\mathbf{r}$ with the size of $1 \times 768$ is learned from \ours{} as seen in Section~\ref{sec:architecture}.

Using Eq.~\eqref{eq:similarity}, we can easily retrieve $K$ vocabulary $\mathcal{V}=\left \{ \hat{v_i}\right\}_{i=1}^{K} $ for $K$ region features $\left \{ \mathbf{r}_i\right\}_{i=1}^{K}$ from $M$ pairs of key-value $(\mathbf{d}, v)$.
In particular, we first obtain the index $j$ of the vocabulary in the external knowledge which has the highest similarity score with the region feature $r_i$.
Then, we assign $v_j$ as the retrieved vocabulary $\hat{v}_i$.
The retrieval process can be summarized as follows:

\begin{equation}
    j = \operatorname*{arg \, max}_j (\operatorname{sim}(\mathbf{r}_i, \mathbf{d}_j)), \quad \hat{v}_i \leftarrow v_j,
\end{equation}
%
where $j \in \left [ 0, M \right )$. 
Because an image may contain multiple identical objects, there is no requirement that each vocabulary is distinct from the others.
Training the retriever is now to find the optimal bipartite match between the retrieved vocabulary and the ground truth objects.
The following is the description of the retriever's training loss function.

Let us denote $\mathcal{Y}=\left \{ y_i \right\}_{i=1}^{N}$ be a set of $N$ ground truth objects.
Assuming that $N < K$, we expand $\mathcal{Y}$ as a set of size $K$ padded with $\varnothing$ (no object) as in~\cite{nicolas2020detr,kim2021hotr}.
However, at this time, the expanded $\mathcal{Y}$ solely consists of seen and no objects, preventing the model from learning to retrieve novel objects.
We thus introduce a simple yet effective modification that replaces 15\% of the number of $\varnothing$ with randomly selected vocabulary from the external knowledge.
As a result, our expanded $\mathcal{Y}$ composes of ground truth objects, no objects and novel vocabulary (objects) (now $|\mathcal{Y}| = |\mathcal{V}| = K$).
Following~\cite{nicolas2020detr,kim2021hotr}, we first find a bipartite matching between $\mathcal{Y}$ and $\mathcal{V}$ by searching for a permutation of $K$ elements $\sigma \in \mathfrak{S}_{N}$ with the lowest cost:
\begin{equation}
    \hat{\sigma} = \operatorname*{arg\, min}_{\sigma \in \mathfrak{S}_{N}} \sum_{i=1}^{K}\mathcal{C}_{\mathrm{match}}(y_i, \hat{v}_{\sigma(i)}),
    \label{eq:bipartite matching}
\end{equation}
%
where $\mathcal{C}_{\mathrm{match}}(y_i, \hat{v}_{\sigma(i)})$ is a pair-wise \textit{matching cost} between $y_i$ and a retrieved vocabulary with index $\sigma(i)$:

\begin{equation}
    \mathcal{C}_{\mathrm{match}}(y_i, \hat{v}_{\sigma(i)})=\mathds{-1}_{y_i \neq \varnothing}\operatorname{sim}(\mathbf{y}_i,\hat{\mathbf{v}}_{\sigma(i)}),
\end{equation}

\noindent
where $\operatorname{sim}(\cdot,\cdot)$ is the same with Eq.~\eqref{eq:similarity}, $\mathbf{y}_i = \operatorname{BERT}(y_i)$, and $\hat{\mathbf{v}}_{\sigma(i)} = \operatorname{BERT}(\hat{v}_{\sigma(i)})$.

Finally, we employ \textit{Hungarian loss} to compute loss for all above pairs matched:
\begin{equation}
    \mathcal{L}_{\mathrm{H}}(\mathcal{Y},\mathcal{V}) = \sum_{i=1}^{K}-\log\operatorname{sim}(\mathbf{y}_i,\hat{\mathbf{v}}_{\hat{\sigma}(i)}),
    \label{eq:hungarian loss}
\end{equation}
%
where $\hat{\sigma}$ is the optimal matching solution in Eq.~\eqref{eq:bipartite matching}.
Like~\cite{nicolas2020detr}, we down-weight the $\log$ term by a factor of 10 when $y_i = \varnothing$ to avoid class imbalance.

\subsection{\ours{} architecture} \label{sec:architecture}

Fig.~\ref{fig:framework} depicts the overall \ours{} architecture which is surprisingly simple, consisting of \textit{only one} multi-layer transformers model in which the parameters are shared between two steps: (1) Vocabulary retrieval and (2) Caption generation.
Like~\cite{li2020oscar,hu2021vivo,zhang2021vinvl}, we employ a pre-trained BERT model~\cite{devlin-etal-2019-bert} to implement our model because BERT is trained on a large corpus of sentences, resulting in a better understanding of grammar and sentence structure.
Following~\cite{li2020oscar,zhang2021vinvl}, we feed our model a concatenated-sequence of words---object tags---ROIs.
This is because, as discussed in~\cite{li2020oscar,zhang2021vinvl}, the concatenated sequence allows for better alignment of vision and language. 
We do not pre-train our model on a large text-image corpus
to eliminate the model's ability to bias to novel objects.
Note that our model only receives ROIs (visual regions) as its input at the testing time.

\noindent
\textbf{Pre-processing.}
For a sequence of words $\left \{ w_i\right\}_{i=1}^{L}$ (i.e., ground truth sentence), we randomly masked out 15\% words of the sequence (maximum 3 words) with a special token $\mathtt{[MASK]}$.
For a given image, we use Faster RCNN~\cite{ren2015faster} trained on the COCO dataset to extract a set of $N$ object tags and a set of $K$ ROIs.
We remark that $N$ tags are used as ground truth objects $\mathcal{Y}=\left \{ y_i\right\}_{i=1}^{N}$ in our vocabulary retrieval as described in Section~\ref{sec:retrieval}.
Each ROI $\mathbf{f}_i$, in contrast, is a vector with the size of $1 \times 2054$ which includes feature $1 \times 2048$ (noticing that this is not region feature used in retrieval step), and the region position $1 \times 6$ (the coordinates of top-left and bottom-right corners, height, and width).

\noindent
\textbf{Step 1: Vocabulary retrieval.}
This step implements the knowledge-based vocabulary retrieval described in Section~\ref{sec:retrieval}.
To ensure that we can retrieve the novel objects added at inference time, we enforce our model as a neural retriever~\cite{guu2020realm,karpukhin2020dense,Lewis2020RAG}, allowing the model to optimize the similarity of image features and external knowledge while training.
Consequently, we only need to update the external knowledge when new objects appear; retraining the whole model is not required.
To this end, we need to optimize both region feature $\mathbf{r}$ and embeddings $\mathbf{d}$ (i.e., training knowledge encoder $\operatorname{BERT}(\cdot)$).
However, because updating $\operatorname{BERT}(\cdot)$ during training is costly and our external knowledge is sufficiently smaller than that used in NLP tasks, we instead fix the parameter of knowledge encoder while training the region feature encoder, similarly to~\cite{Lewis2020RAG}.

We begin with encoding each word $w_i$ and object tag $y_i$ into a vector $1 \times 768$ using an embeddings layer.
Simultaneously, for each ROI $\mathbf{f}_i$, we use another embeddings layer to reduce its size from $1 \times 2054$ to $1 \times 768$.
The vectors are then concatenated with two special tokens $\mathtt{[SEP]}$ to distinguish their functions.
More precisely, we have $\left \{\mathbf{w}_1,\ldots,\mathbf{w}_L,\mathtt{[SEP]},\mathbf{y}_1,\ldots,\mathbf{y}_N,\mathtt{[SEP]},\mathbf{f}_1,\ldots,\mathbf{f}_K  \right \}$ as an input.
After feeding the above input to the model, we obtain $K$ region features $\left \{ \mathbf{r}_i\right\}_{i=1}^{K}$.
Finally, we perform the retrieval using $K$ region features and the external knowledge, returning retrieved vocabulary $\mathcal{V}$.

\noindent
\textbf{Step 2: Caption generation.}
This step aims to generate a caption from the given image and the vocabulary retrieved in the previous step.
This step, like the vocabulary retrieval step, takes as its input a concatenated-sequence of words---object tags---ROIs.
However, we use our retrieved vocabulary to replace the object tags detected by Faster RCNN~\cite{ren2015faster}.
As a result, the input is changed to $\left \{\mathbf{w}_1,\ldots,\mathbf{w}_L,\mathtt{[SEP]},\hat{\mathbf{v}}_1,\ldots,\hat{\mathbf{v}}_K,\mathtt{[SEP]},\mathbf{f}_1,\ldots,\mathbf{f}_K  \right \}$.

\noindent
\textbf{Inference.}
Our model's input is an image that has been pre-processed to obtain ROIs.
It automates the steps of vocabulary retrieval and caption generation in an end-to-end manner.
To avoid significant changes in input between training and testing, we create $(L+N)$ $\mathtt{[MASK]}$ as pseudo words.

\subsection{Loss function}
We define our loss function as: $\mathcal{L} = \mathcal{L}_{\mathrm{H}} + \mathcal{L}_{\mathrm{G}}$.
$\mathcal{L}_{\mathrm{H}}$ is used for the vocabulary retrieval step as defined in Eq.~\ref{eq:hungarian loss} while $\mathcal{L}_{\mathrm{G}}$ works for the caption generation step. 
$\mathcal{L}_{\mathrm{G}}$ is defined as follows:
\begin{equation}
   \mathcal{L}_{\mathrm{G}} = \begin{cases}
\operatorname{cross\_entropy}(S, S_{\rm GT}) & \text{at 1st training stage} \\ 
 \operatorname{CIDEr}(S) + \alpha \times C & \text{at 2nd training stage}
\end{cases},
\label{eq:caption loss}
\end{equation}
%
where $S$, $S_{\rm GT}$ are generated and ground truth captions.
$\operatorname{cross\_entropy}(\cdot, \cdot)$ is the cross entropy loss function.
Meanwhile, $\operatorname{CIDEr}(\cdot)$ plays as SCST function~\cite{Rennie2017SCST} to improve caption quality (e.g., grammar, structure) and $C$ provides an additive reward for each retrieved vocabulary appearing in the caption.
A small $\alpha$ balances the two terms.

\section{Experiments}

\begin{figure*}[tb]
	\centering
	\includegraphics[width=1\linewidth]{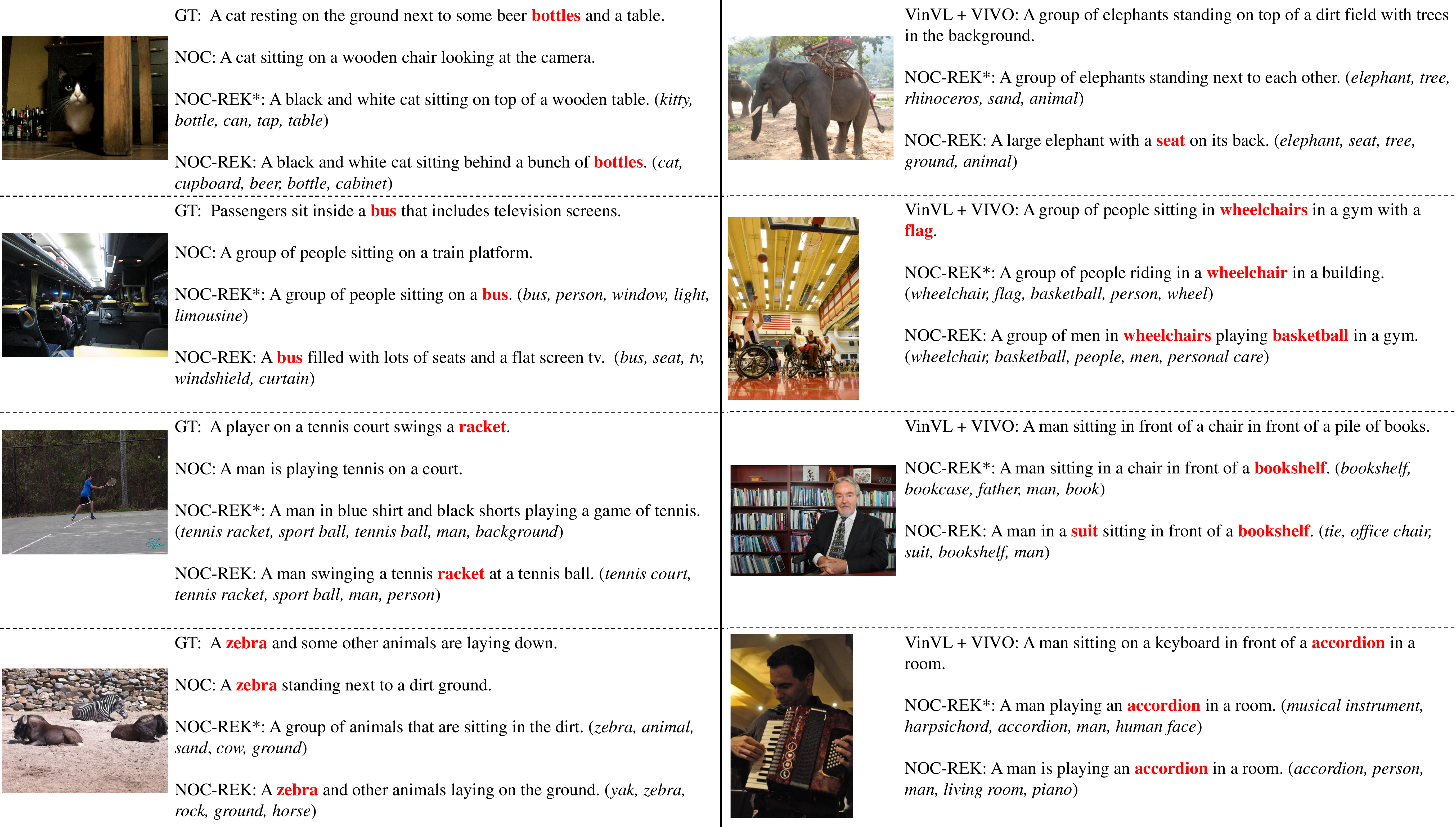}
	\caption{Examples of generated captions by compared methods on held-out COCO (left) and Nocaps (right). We show the ground-truth captions (GT) on held-out COCO for reference. 
	On held-out COCO, NOC~\cite{venugopalan17noc} usually fails to generate captions with novel objects (first three examples) or generate caption with not related to image's context (fourth example).
	On Nocaps, VinVL+VIVO~\cite{zhang2021vinvl,hu2021vivo} sometimes cannot include the novel objects in the captions (first and third examples) or generates weird caption (fourth example).
	Our \ours{}, on the other hand, successfully generates correct, fluent, and coherent captions with novel objects. 
	Words in parentheses are top-5 retrieved vocabulary by our method that are reasonably related to objects in image. \red{Red} texts indicate novel objects in the captions.} 
	\label{fig:captions}
	\vspace*{-0.25\baselineskip}
\end{figure*}

\subsection{Implementation and training details}

\ours{} was built with PyTorch, and we used a pre-trained BERT-base model from Huggingfaces~\cite{huggingfaces} for parameters initialization.
Our model was trained in an end-to-end manner with two-stage optimization using AdamW, with $\mathcal{L}_{\mathrm{H}}$ used in both stages and $\mathcal{L}_{\mathrm{G}}$ changed with respect to Eq.~\eqref{eq:caption loss}.
We set the learning rate to 3e-5 and the batch size to 128 for 30 epochs during the first training stage.
The learning rate, batch size, and epoch at the second training stage, in contrast, are 8e-7, 6, and 25, respectively.
On a PC with two GTX-3090 GPUs, our model takes 8 days to train.
\ours{}* denotes our method after the first training stage, and \ours{} denotes the fully-trained model.

We set the length of the word sequence $L=35$, the number of ground-truth objects $N=20$, the number of ROIs $K=50$, and $\alpha=0.3$.
During training, we use seen objects in held-out COCO~\cite{hendricks16dcc} as the external knowledge ($M=72+1=73$).
During inference, novel objects in held-out COCO~\cite{hendricks16dcc} and Nocaps~\cite{Agrawal2019nocaps} will be added to the knowledge ($M=600+1=601$).
We retain top-5 retrieved vocabulary with the highest similarity score at inference time (i.e., 5 objects per image).
We use CBS~\cite{Anderson2017CBS} with the beam size of 5 to generate caption.

\subsection{Compared methods and evaluation metrics}

\noindent
\textbf{Dataset and compared methods.}
We evaluate \ours{} on held-out COCO~\cite{hendricks16dcc} and Nocaps~\cite{Agrawal2019nocaps} datasets.
Held-out COCO~\cite{hendricks16dcc} is made by dividing the original COCO~\cite{lin2014coco} into known classes and 8 novel classes which include \textit{bottle, bus, couch, microwave, pizza, racket, suitcase, zebra}.
On the other hand, Nocaps~\cite{Agrawal2019nocaps} is the main challenging dataset for novel object captioning task, which is constructed from 513 classes out of 600 classes of OpenImage.
In particular, it consists of 119 classes in COCO dataset that are \textit{in-domain}, 394 classes not in COCO dataset are \textit{out-domain}, and the images that includes both in-domain and out-domain are regarded as \textit{near-domain}.
We validate our method using Nocaps validation and test sets.

We compare \ours{} with state-of-the-art methods: UpDown~\cite{Anderson2017up-down}, DCC~\cite{hendricks16dcc}, NOC~\cite{venugopalan17noc}, NBT~\cite{Lu2018nbt}, DNOC~\cite{wu2018dnoc}, ZSC~\cite{Berkan2019zsc}, OVE~\cite{Mikihiro2020ove}, ANOC~\cite{xianyu2021anoc}, Oscar~\cite{li2020oscar}, VIVO~\cite{hu2021vivo}, and VinVL~\cite{zhang2021vinvl} in Sections~\ref{sec:heldout} and~\ref{sec:nocaps}.
We note that all compared scores are from published results, whereas for qualitative comparison, we use publicly available caption results (NOC~\cite{noc}) and pre-trained models (VinVL+VIVO~\cite{vinvl}).
In Section~\ref{sec:analysis}, we go over the performance of two variants of our method: NOC-REK and NOC-REK*.

\noindent
\textbf{Evaluation metrics.}
We primarily report CIDEr~\cite{Vedantam2015CIDEr} and SPICE~\cite{Anderson2016SPICE} language scores in comparisons on both held-out COCO and Nocaps datasets, drawing on prior work.
We also report object detection results on held-out COCO using F1-score and language score METEOR~\cite{Lavie2007METEOR}.

\subsection{Results on held-out COCO dataset} \label{sec:heldout}

\begin{table*}[tb]
{\centering
\caption{Quantitative comparison against other methods on held-out COCO dataset. We report F1-score for each novel class (2nd - 9th columns), average F1-score on all novel classes (10th column), and language scores (11th - 13rd columns)$^a$. Higher score is better.} 
	\vspace*{-0.75\baselineskip}
\label{tab:heldout}
\resizebox{\linewidth}{!}{
\begin{tabular}{l|cccccccc|cccc}
\toprule
\multirow{2}{*}{Method} & \multicolumn{8}{c|}{F1-score} & Avg. F1-score & SPICE & METEOR & CIDEr \\
 & bottle & bus & couch & microwave & pizza & racket & suitcase & zebra \\
\hline
DCC~\cite{hendricks16dcc} & 4.6 & 29.8 & 45.9 & 28.1 & 64.6 & 52.2 & 13.2 & 79.9 & 39.8 & 13.4 & 21.0 & 59.1 \\
NOC~\cite{venugopalan17noc} & 17.8 & 68.8 & 25.6 & 24.7 & 69.3 & 55.3 & 39.9 & \red{89.0} & 48.8 & -- & 21.4 & -- \\
NBT~\cite{Lu2018nbt} & 7.1 & 73.7 & 34.4 & 61.9 & 59.9 & 20.2 & 42.3 & \blue{88.5} & 48.5 & 15.7 & 22.8 & 77.0 \\
DNOC~\cite{wu2018dnoc} & 33.0 & \red{76.9} & \red{54.0} & 46.6 & 75.8 & 33.0 & 59.5 & 84.6 & 57.9 & -- & 21.6 & -- \\
ZSC~\cite{Berkan2019zsc} & 2.4 & 75.2 & 26.6 & 24.6 & 29.8 & 3.6 & 0.6 & 75.4 & 29.8 & 14.2 & 21.9 & -- \\
OVE~\cite{Mikihiro2020ove} & -- & -- & -- & -- & -- & -- & -- & -- & -- & 16.9 & 22.6 & 80.8 \\
ANOC~\cite{xianyu2021anoc} & -- & -- & -- & -- & -- & -- & -- & -- & 64.3 & 18.2 & 25.2 & 94.7\\
\hline
\rowcolor{lightgray}
\ours{}* & \red{39.3} & 76.2 & 47.4 & \red{62.0} & \red{79.5} & \red{78.6} & \red{72.6} & 85.8 & \red{67.7} & \red{23.4} & \red{30.3} & \red{109.3} \\
\rowcolor{lightgray}
\ours{} & \blue{59.4} & \blue{79.3} & \blue{67.8} & \blue{73.4} & \blue{82.1} & \blue{81.1} & \blue{79.9} & 87.2 & \blue{76.3} & \blue{26.9} & \blue{32.8} & \blue{138.4} \\
\hline 
$\Delta$ & 26.4$\uparrow$ & 2.4$\uparrow$ & 13.8$\uparrow$ & 11.5$\uparrow$ & 12.8$\uparrow$ & 25.8$\uparrow$ & 20.4$\uparrow$ & 1.8$\downarrow$ & 12.0$\uparrow$ & 8.7$\uparrow$ & 7.6$\uparrow$ & 43.7$\uparrow$ \\
\bottomrule
\end{tabular}
}
}
\footnotesize{$^a$For all the tables in this paper, \blue{Blue} indicates the best results among compared method (not applicable to results by human), \red{Red} indicates the second best results, gray background indicates results obtained by our method, and $\Delta$ indicates the improvement over state-of-the-art methods.}
\vspace*{-1.5\baselineskip}
\end{table*}

\begin{table}[tb]
\centering
\caption{Caption generation evaluation using CIDEr on held-out COCO dataset. We report the score for each class (higher score is better). Our method significantly outperforms OVE~\cite{Mikihiro2020ove}.} 
	\vspace*{-0.75\baselineskip}
\label{tab:cider_on_each_class_heldout}
\resizebox{\linewidth}{!}{
\begin{tabular}{l|cccccccc}
\toprule
Method & bottle & bus & couch & microwave & pizza & racket & suitcase & zebra \\
\hline
OVE~\cite{Mikihiro2020ove} & 67.2 & 79.7 & \red{97.4} & \red{98.4} & 86.5 & \blue{103.9} & 56.2 & \red{84.1}  \\
\hline
\rowcolor{lightgray}
\ours{}* & \red{102.1} & \red{81.0} & 89.2 & 71.9 & \red{87.6} & 60.0 & \red{84.3} & 56.2 \\
\rowcolor{lightgray}
\ours{} & \blue{141.1} & \blue{118.9} & \blue{124.3} & \blue{108.0} & \blue{110.0} & \red{89.1} & \blue{116.9} & \blue{99.0} \\
\hline
$\Delta$ & 73.9$\uparrow$ & 39.2$\uparrow$ & 26.9$\uparrow$ & 9.6$\uparrow$ & 23.5$\uparrow$ & 14.8$\downarrow$ & 60.7$\uparrow$ & 14.9$\uparrow$ \\
\bottomrule
\end{tabular}
}
\vspace*{-1.25\baselineskip}
\end{table}

\noindent
\textbf{Qualitative evaluation.}
Fig.~\ref{fig:captions} (left) shows randomly picked-up captions generated by our method and NOC~\cite{venugopalan17noc}.
We can see that our method successfully retrieves novel objects and includes some of them in the captions in a sensible fashion.
Meanwhile, NOC~\cite{venugopalan17noc} usually fails to generate a caption with a novel object (see first three examples).
Furthermore, NOC~\cite{venugopalan17noc} generates a pretty weird caption (fourth example) that is unrelated to the image's context. 
This is because NOC~\cite{venugopalan17noc} uses text-corpus as its external knowledge, resulting in language biases (e.g., a \textit{zebra} is usually described with the word \textit{standing} rather than \textit{laying} in the text-corpus).
On the other hand, our method is free of such biases thanks to our usage of object definitions as external knowledge.
We also see that our top-5 retrieved vocabulary are reasonable, reflecting objects in the given images.
Thanks to end-to-end training, our model can remove less relevant vocabulary in the captions.
In more detail, our captions are correct, fluent, and coherent, just like ground-truth captions, proving that our usage of a pre-trained BERT model is reasonable.

\noindent
\textbf{Quantitative evaluation.}
We first evaluate whether a novel object appears in the generated caption using the F1-score.
The per-object F1-score and average F1-score of all the compared methods are shown in Table~\ref{tab:heldout} (2nd -- 10th columns).
Notably, despite the lower cost of updating novel object information, our method yields significantly higher F1-scores than the other methods (an improvement of 12\% in average F1-score).
This indicates that our method successfully retrieves the novel object from the external knowledge (Fig.~\ref{fig:captions}), and includes those objects in the caption.

Next, we quantitatively evaluate the quality of generated captions using SPICE, METEOR, and CIDEr scores (Table~\ref{tab:heldout}, 11th -- 13rd columns).
Because our method can include novel objects in captions, it is not surprising that we outperform the other methods by a considerable margin.
Together with the F1-score, we can conclude that our method outperforms the other methods marginally.
More importantly, these observations strongly support the advantage of using an end-to-end model in this task.

Finally, as shown in Table~\ref{tab:cider_on_each_class_heldout}, we look further into the CIDEr score for each class.
While our method achieves a higher CIDEr score for almost all classes, we notice that it underperforms on the \textit{racket} class.
This is due to a mismatch between the word forms \textit{racket} and \textit{racquet} in our generated captions and the ground truth captions.
We believe that this is a minor issue that does not detract from our superiority over the other methods in general.

\subsection{Results on Nocaps dataset} \label{sec:nocaps}

\noindent
\textbf{Qualitative evaluation.}
Fig.~\ref{fig:captions} (right) shows examples of generated captions obtained by our method and VinVL+VIVO~\cite{zhang2021vinvl,hu2021vivo}.
VinVL+VIVO occasionally cannot generate captions consisting of novel objects (first and third examples).
In addition, VinVL+VIVO also generates a weird caption (fourth example).
In contrast, our method effectively allows the appearance of novel objects in the generated captions.
This advantage can be attributed to our usage of the reward for encouraging retrieved vocabulary to appear in the caption.
Note that top-5 retrieved vocabulary by our method are reasonable though some are not correct because Nocaps is more challenging than held-out COCO.

\begin{table*}[tb]
{\centering
\caption{Caption generation evaluation using SPICE and CIDEr on the Nocaps validation and test sets. We achieve the best scores for in-domain, near-domain, out-domain and Overall (excepting for CIDEr on near-domain of test set). Notably, the captions by our method are better than those by human in most cases. We note that our results on test set are better than those by other methods which are publicly submitted to Nocaps leader-board$^b$. Higher score is better.} 
	\vspace*{-0.75\baselineskip}
\label{tab:nocaps}
\resizebox{\linewidth}{!}{
\begin{tabular}{l|cc|cc|cc|cc|cc|cc|cc|cc}
\toprule
\multirow{3}{*}{Method} & \multicolumn{8}{c|}{Validation set} & \multicolumn{8}{c}{Test set} \\
& \multicolumn{2}{c|}{in-domain} & \multicolumn{2}{c|}{near-domain} & \multicolumn{2}{c|}{out-domain} & \multicolumn{2}{c|}{Overall} & \multicolumn{2}{c|}{in-domain} & \multicolumn{2}{c|}{near-domain} & \multicolumn{2}{c|}{out-domain} & \multicolumn{2}{c}{Overall}\\
& CIDEr & SPICE & CIDEr & SPICE & CIDEr & SPICE & CIDEr & SPICE & CIDEr & SPICE & CIDEr & SPICE & CIDEr & SPICE & CIDEr & SPICE\\
\hline
UpDown~\cite{Anderson2017up-down} & 78.1 & 11.6 & 57.7 & 10.3 & 31.3 & 8.3 & 55.3 & 10.1 & 76.0 & 11.8 & 74.2 & 11.5 & 66.7 & 9.7 & 73.1 & 11.2  \\
OVE~\cite{Mikihiro2020ove} & 79.5 & 11.5 & 74.7 & 11.1 & 78.2 & 10.7 & 76.1 & 11.1 & -- & -- & -- & -- & -- & -- & -- & --\\
ANOC~\cite{xianyu2021anoc} & 86.1 & 12.0 & 80.7 & 11.9 & 73.7 & 10.1 & 80.1 & 11.6 & 85.8 & 12.4 & 79.7 & 11.8 & 68.5 & 10.0 & 78.5 & 11.6 \\
Oscar~\cite{li2020oscar} & 83.4 & 12.0 & 81.6 & 12.0 & 77.6 & 10.6 & 81.1 & 11.7 & 81.3 & 11.9 & 79.6 & 11.9 & 73.6 & 10.6 & 78.8 & 11.7 \\
VIVO~\cite{hu2021vivo} & 92.2 & 12.9 & 87.8 & 12.6 & \red{87.5} & 11.5 & 88.3 & 12.4 & 89.0 & 12.9 & 87.8 & 12.6 & 80.1 & 11.1 & 86.6 & 12.4 \\
VinVL~\cite{zhang2021vinvl} & 96.8 & 13.5 & 90.7 & 13.1 & 87.4 & 11.6 & 90.9 & 12.8 & 93.8 & 13.3 & 89.0 & 12.8 & 66.1 & 10.9 & 85.5 & 12.5 \\
VinVL + VIVO~\cite{zhang2021vinvl,hu2021vivo} & \red{103.7} & \red{13.7} & \red{95.6} & \red{13.4} & 83.8 & \red{11.9} & \red{94.3} & \red{13.1} & \red{98.0} & \red{13.6} & \red{95.2} & \red{13.4} & \blue{78.0} & \red{11.5} & \red{92.5} & \red{13.1} \\
\hline
\rowcolor{lightgray}
\ours{}* & 88.3 & 12.5 & 83.0 & 12.2 & 79.0 & 10.8 & 82.9 & 12.0 & 89.2 & 13.3 & 86.6 & 13.1 & 70.2 & 11.1 & 84.0 & 12.7 \\
\rowcolor{lightgray}
\ours{} & \blue{104.7} & \blue{14.8} & \blue{100.2} & \blue{14.1} & \blue{100.7} & \blue{13.0} & \blue{100.9} & \blue{14.0} &  \blue{100.0} & \blue{14.1} & \blue{95.7} & \blue{13.6} & \red{77.4} & \blue{11.6} & \blue{93.0} & \blue{13.4} \\
\hline
$\Delta$ & 1.0$\uparrow$ & 1.1$\uparrow$ & 4.6$\uparrow$ & 0.7$\uparrow$ & 13.2$\uparrow$ & 1.1$\uparrow$ & 6.6$\uparrow$ & 0.9$\uparrow$ & 2.0$\uparrow$ & 0.5$\uparrow$ & 0.5$\uparrow$ & 0.2$\uparrow$ & 0.6$\downarrow$ & 0.1$\uparrow$ & 0.5$\uparrow$ & 0.3$\uparrow$ \\
\hline
Human~\cite{Agrawal2019nocaps} & 84.4 & 14.3 & 85.0 & 14.3 & 95.7 & 14.0 & 87.1 & 14.2 & 80.6 & 15.0 & 84.6 & 14.7 & 91.6 & 14.2 & 85.3 & 14.6 \\
\bottomrule
\end{tabular}
}
}
\footnotesize{$^b$\url{https://eval.ai/web/challenges/challenge-page/355/leaderboard/1011}}
\vspace*{-1.25\baselineskip}
\end{table*}

\noindent
\textbf{Quantitative evaluation.}
Table~\ref{tab:nocaps} quantitatively compares caption generation on Nocaps validation and test sets using SPICE and CIDEr.
Except for CIDEr on the out-domain of the test set, we see that our method outperforms the others by a large margin.
The performance of VinVL+VIVO is comparable to ours because they use a visual vocabulary pre-trained model that covers all the objects in Nocaps.
However, without VIVO, the VinVL itself cannot defeat our method.
Therefore, we conclude that \ours{} consistently works for all domains, particularly the out-domain, which contains all novel objects.
Nonetheless, despite having higher scores than humans in most cases, our method performs noticeably worse than humans on the test set's out-domain.
This demonstrates the task's difficulty, as well as the requirement for further improvement.
Note that our method surpasses VIVO~\cite{hu2021vivo} and VinVL~\cite{zhang2021vinvl} on the Nocaps online leaderboard (accessed on Oct. 9, 2021).

\subsection{Detailed analysis} \label{sec:analysis}

\begin{figure}[tb]
	\centering
	\includegraphics[width=1\linewidth]{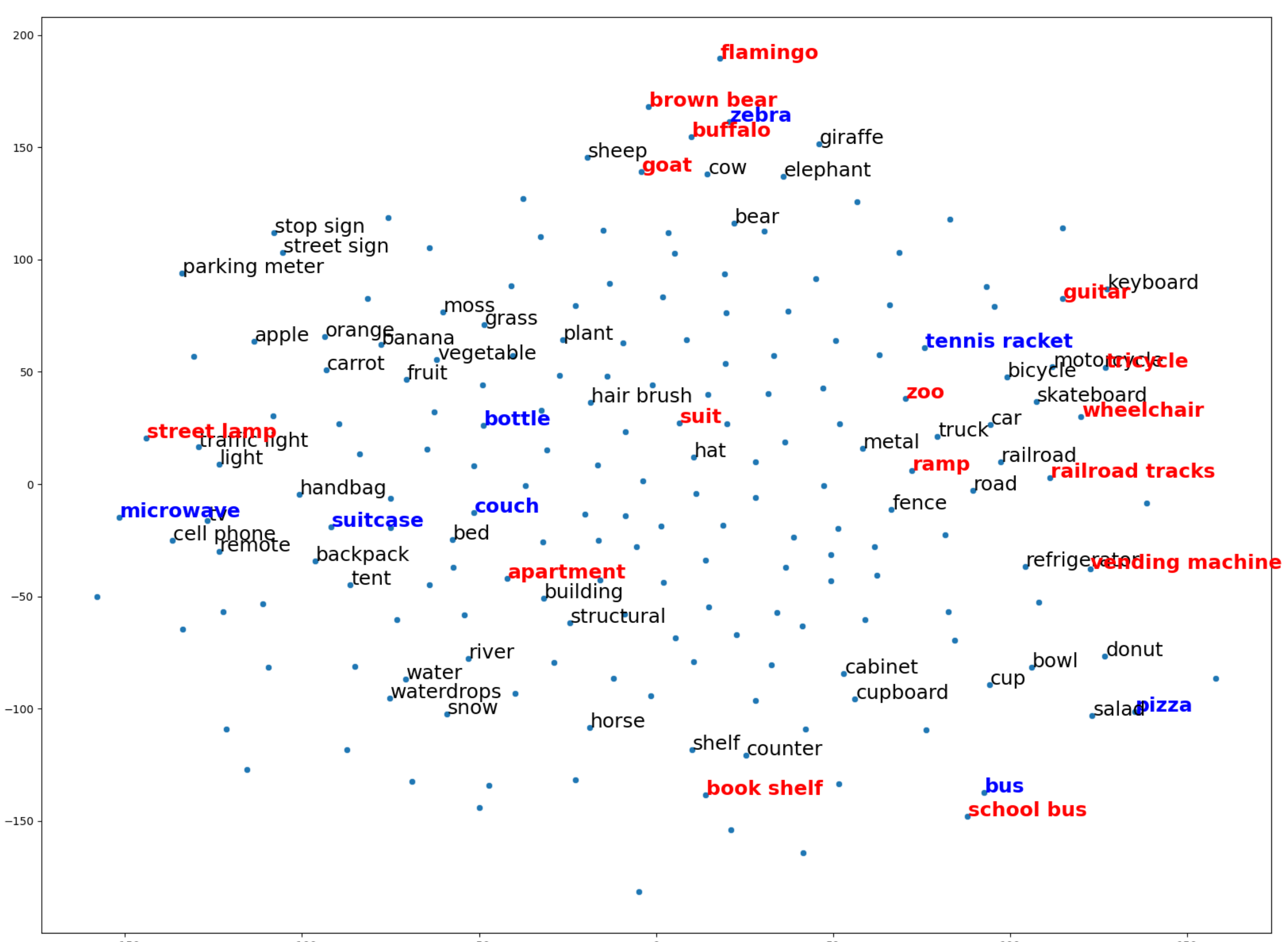}
    \caption{Visualization of external knowledge using t-SNE. We see that the related vocabulary (we use objects from the seen classes of held-out COCO dataset) fall in the same cluster (black text). When we add more objects from novel classes of held-out COCO dataset (blue text) and Nocaps dataset (red text), the novel vocabulary are located at the appropriate cluster.} 
	\label{fig:knowledge_visualization}
	\vspace*{-1.5\baselineskip}
\end{figure}

\noindent
\textbf{Visualization of external knowledge.}
We visually investigate our collected knowledge to better understand why our method can retrieve the vocabulary added during inference time.
We use t-SNE~\cite{tsne} to reduce the dimension of each $\mathbf{d}$ from $1 \times 768$ to $1 \times 2$, and then plot the reduced $\mathbf{d}$ into 2-D plan (Fig.~\ref{fig:knowledge_visualization}).
Note that we display the vocabulary for corresponding reduced $\mathbf{d}$ in Fig.~\ref{fig:knowledge_visualization}.
For the sake of simplicity, we show some clusters at random here, while a full visualization is provided in the supplementary.
Fig.~\ref{fig:knowledge_visualization} shows that our external knowledge not only adequately clusters the vocabulary (black texts) but also locates new vocabulary (red and blue texts) into appropriate clusters.
We optimize the similarity of image features and embeddings in external knowledge, which means that the image feature is also located in the relevant vocabulary cluster.
The image feature of a novel object, on the other hand, would be similar to some seen objects to some extent~\cite{Demirel2018ZeroShotOD}.
When computing its similarity to the external knowledge, the novel object will intuitively fall into the same cluster as its related objects.
Consequently, we can pick up on the new vocabulary properly.
In fact, the model sometimes cannot retrieve a reasonable vocabulary.
In the fourth example on Nocaps (Fig.~\ref{fig:captions}, right), for example, our model retrieves both \textit{accordion} and \textit{piano} because their appearances are too similar.

\noindent
\textbf{Ablation study.}
We evaluate the performance of using different loss term in Eq.~\ref{eq:caption loss} by comparing \ours{}* and \ours{}.
As shown in Fig.~\ref{fig:captions}, both \ours{} and \ours{}* are capable of retrieving appropriate vocabulary and generating reasonable captions.
However, while \ours{} can include more novel objects in the captions, \ours{}* cannot.
When we quantitatively evaluate caption generation, \ours{}* performs worse than \ours{} (see Tables~\ref{tab:heldout},~\ref{tab:cider_on_each_class_heldout} and~\ref{tab:nocaps}).
The following is an explanation for the deterioration.
We use cross-entropy loss between generated and ground truth captions in the first training stage to enforce the model to generate captions similar to those in the training dataset.
Because the dataset contains no novel objects, it is difficult for \ours{}* to describe them.
In contrast, the CIDEr optimizer loss is not strictly related to ground truth captions at the second training stage, giving the model a better chance of describing novel objects.
Furthermore, because we encourage more novel objects to appear in the second training stage, \ours{} performs better than \ours{}*.
Note that both \ours{}* and \ours{} produce results that are at least comparable to the other methods.

We do not investigate the ablated model where the vocabulary retrieval and caption generation are trained independently because, as discussed in~\cite{venugopalan17noc}, it raises the difficulty to include novel objects in the captions.
Indeed, most of our compared methods perform far worse than our method, highlighting the drawbacks of independently training.

\begin{table}[tb]
\centering
\caption{Impact of the size of the external knowledge on caption generation evaluation. Changes in the size of external knowledge result in changes in performance. Higher score is better.} 
	\vspace*{-0.5\baselineskip}
\label{tab:knowledge size}
\resizebox{\linewidth}{!}{
\begin{tabular}{l|cc|cc|cc|cc}
\toprule
\multirow{2}{*}{Size of external knowledge} & \multicolumn{2}{c|}{in-domain} & \multicolumn{2}{c|}{near-domain} & \multicolumn{2}{c|}{out-of-domain} & \multicolumn{2}{c}{Overall}\\
& CIDEr & SPICE & CIDEr & SPICE & CIDEr & SPICE & CIDEr & SPICE \\
\midrule
Full (COCO + Nocaps) & 104.7 & 14.8 & 100.2 & 14.1 & 100.7 & 13.0 & 100.9 & 14.0 \\
\quad --25\% & 90.9 & 12.8 & 85.5 & 12.4 & 83.1 & 11.3 & 85.8 & 12.3 \\
\quad --50\% & 87.8 & 12.5 & 83.5 & 12.3 & 79.2 & 10.9 & 83.0 & 12.0 \\
\quad --75\% & 86.4 & 12.4 & 83.1 & 12.2 & 79.2 & 10.8 & 83.2 & 12.0 \\
\quad --COCO & 86.3 & 12.3 & 82.9 & 12.2 & 79.0 & 10.8 & 82.9 & 12.0 \\
\quad --Nocaps & 88.0 & 12.5 & 83.2 & 12.2 & 77.2 & 10.7 & 82.4 & 11.9 \\
\bottomrule
\end{tabular}
}
\vspace*{-\baselineskip}
\end{table}

\begin{figure}[tb]
	\centering
	\includegraphics[width=1\linewidth]{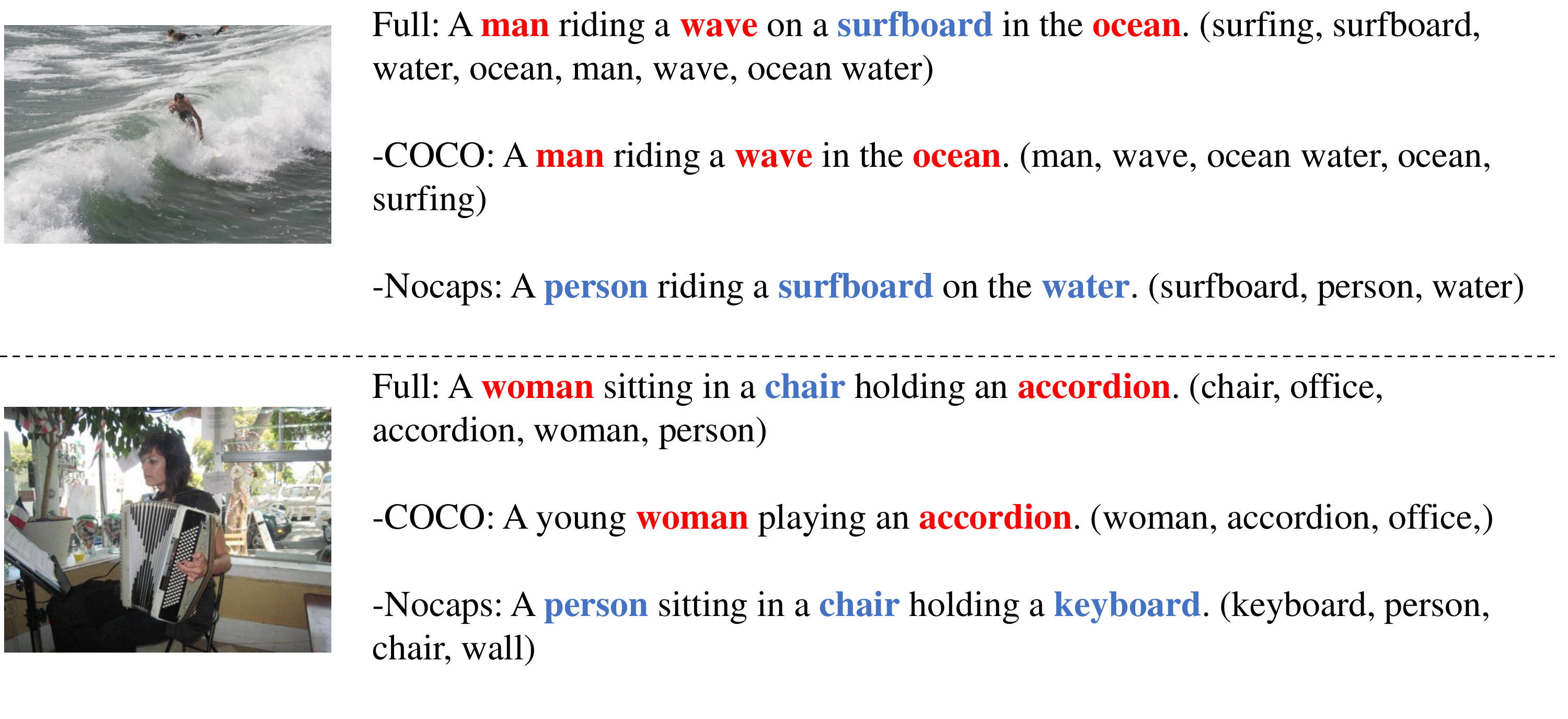}
	\caption{Examples of generated captions using different external knowledge on Nocaps dataset: full knowledge (Full), knowledge without COCO objects (--COCO), knowledge without Nocaps objects (--Nocaps). Words in parentheses are top-5 retrieved vocabulary. \blue{Blue} text is object in COCO, \red{Red} text is object in Nocaps.}
	\label{fig:change_vocabulary}
	\vspace*{-\baselineskip}
\end{figure}

\noindent
\textbf{Impact of the size of the external knowledge.}
We investigate the performance of our model on the Nocaps validation set under various sizes of external knowledge (we cannot use the Nocaps test set because only 5 times submissions are allowed for the test set). 
We use various scenarios to drop vocabulary from a full external knowledge (COCO + Nocaps):
we remove 25\%, 50\%, and 75\% of vocabulary at random.
We also remove any vocabulary found in COCO or Nocaps (out-domain) datasets.
Table~\ref{tab:knowledge size} demonstrates that reducing our vocabulary (25\%, 50\%, and 75\%) leads to degrading the performance.
Moreover, dropping either COCO or Nocaps vocabulary results in a trade-off between in-domain and out-domain performance.
Our model can deal with out-domain but not in-domain without COCO vocabulary and vice versa without Nocaps vocabulary.
These observations also confirm our model's ability to retrieve novel vocabulary from external knowledge and incorporate it into captions.
This experiment also demonstrates how the size of our external knowledge affects our performance, implying that more novel objects are more effective.
Fig.~\ref{fig:change_vocabulary} illustrates that using different vocabulary, we obtain other captions.
We remark that we do not retrain our model in this experiment.

\noindent
\textbf{Limitations.}
First, our model includes a pre-processing step to extract ROIs from a given image, which may not fully explore all potential objects in the image.
To improve the capability of our method, one possible solution is to divide the image into multiple patches to which the transformers can directly attend, as discussed in~\cite{dosovitskiy2020vit}.
Second, the quality of the external knowledge has a significant impact on our method, as seen in Table~\ref{tab:knowledge size} and Fig.~\ref{fig:change_vocabulary}.
As previously stated, it is reasonable to fix the knowledge embeddings as our external knowledge is sufficiently small.
However, in the case of an explosion of novel vocabulary, training both image features and vocabulary embeddings is preferable.
We have left detailed investigations for our future work.

\section{Conclusion}
We streamline the pipeline of novel object captioning by introducing the end-to-end \ours{} model that includes vocabulary retrieval and caption generation steps.
\ours{} learns to retrieve vocabulary from external knowledge and generates captions using shared-parameters transformers.
Our model does not require retraining; instead, it updates the external knowledge whenever new objects become available.
We thoroughly compare our method with SOTAs on held-out COCO and Nocaps datasets, demonstrating significant superiority of \ours{}.

\noindent
\textbf{Acknowledgement.}
This work was supported by the Institute of AI and Beyond of the University of Tokyo, the Next Generation Artificial Intelligence Research Center of the University of Tokyo, and JSPS KAKENHI Grant Number JP19H04166. Duc Minh Vo and Hong Chen are thankful to Yi-Pei Chen for her valuable comments on this work.

{\small
\bibliographystyle{IEEEtran}
\bibliography{references}
}

\clearpage

\appendix

\setcounter{figure}{0}

\renewcommand\thesection{\Roman{section}}
\renewcommand\thesubsection{\Roman{section}.\Alph{subsection}}
\renewcommand\thefigure{\Alph{figure}} 
\renewcommand\thetable{\Alph{table}} 

\twocolumn[\centering{\Large \bf Supplementary Material}]

\vspace*{\baselineskip}

\justifying

\section{Visualization of external knowledge}

\begin{figure*}[tb]
	\centering
	\includegraphics[width=1\linewidth]{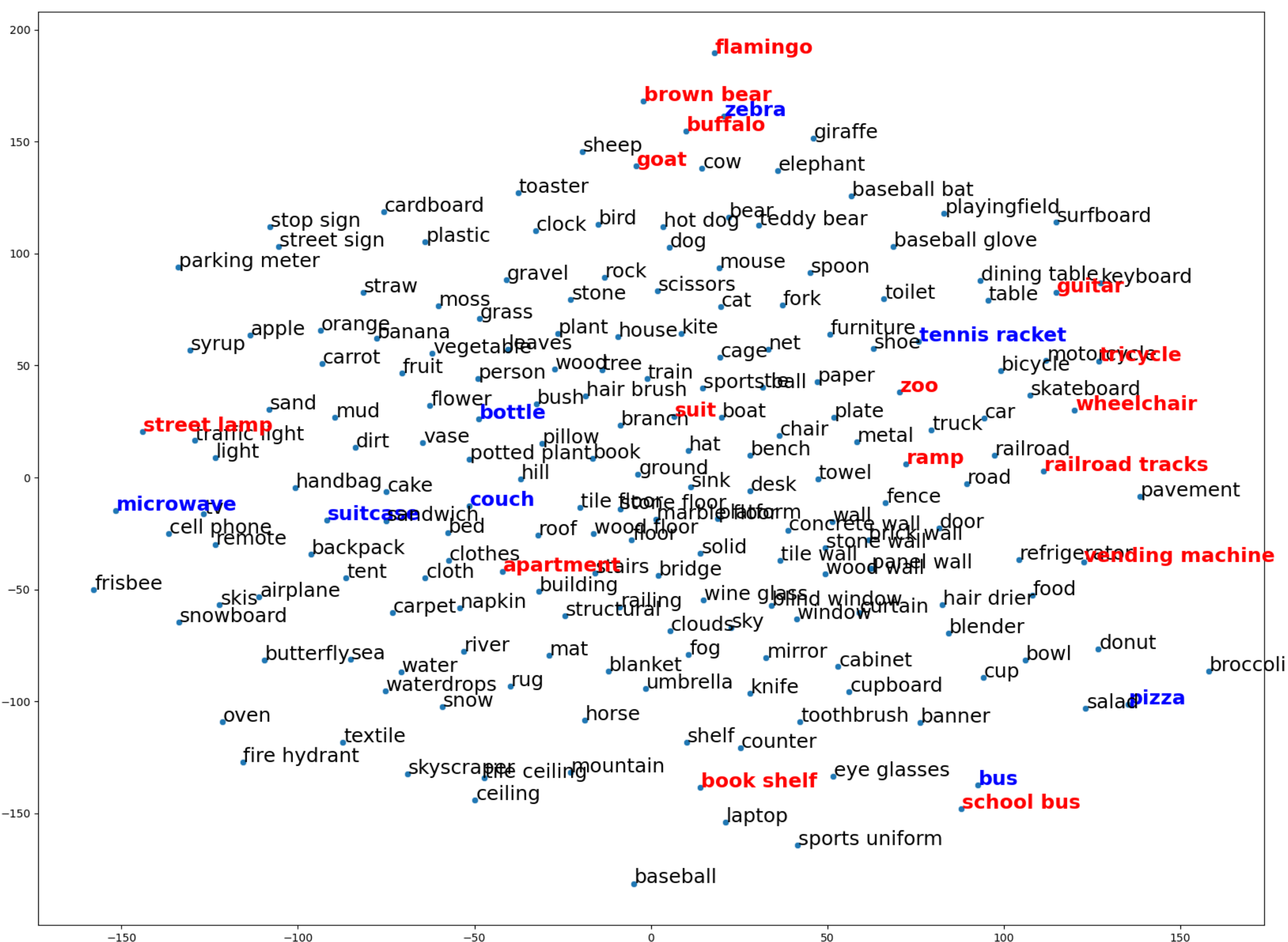}
	\caption{Visualization of external knowledge using t-SNE. We see that the related vocabulary (we use objects from the seen classes of held-out COCO dataset) fall in the same cluster (black text). When we add more objects from novel classes of held-out COCO dataset (blue text) and some classes of Nocaps dataset (red text), the novel vocabulary are located at the appropriate cluster.}
	\label{fig:full_knowledge_visualization}
\end{figure*}

We show full visualization of our collected external knowledge in Fig.~\ref{fig:full_knowledge_visualization}.
We see that our external knowledge not only adequately clusters the vocabulary (black texts), but also locates new vocabulary (red and blue texts) into appropriate clusters.

\section{More examples}

\begin{figure*}[tb]
	\centering
	\includegraphics[width=1\linewidth]{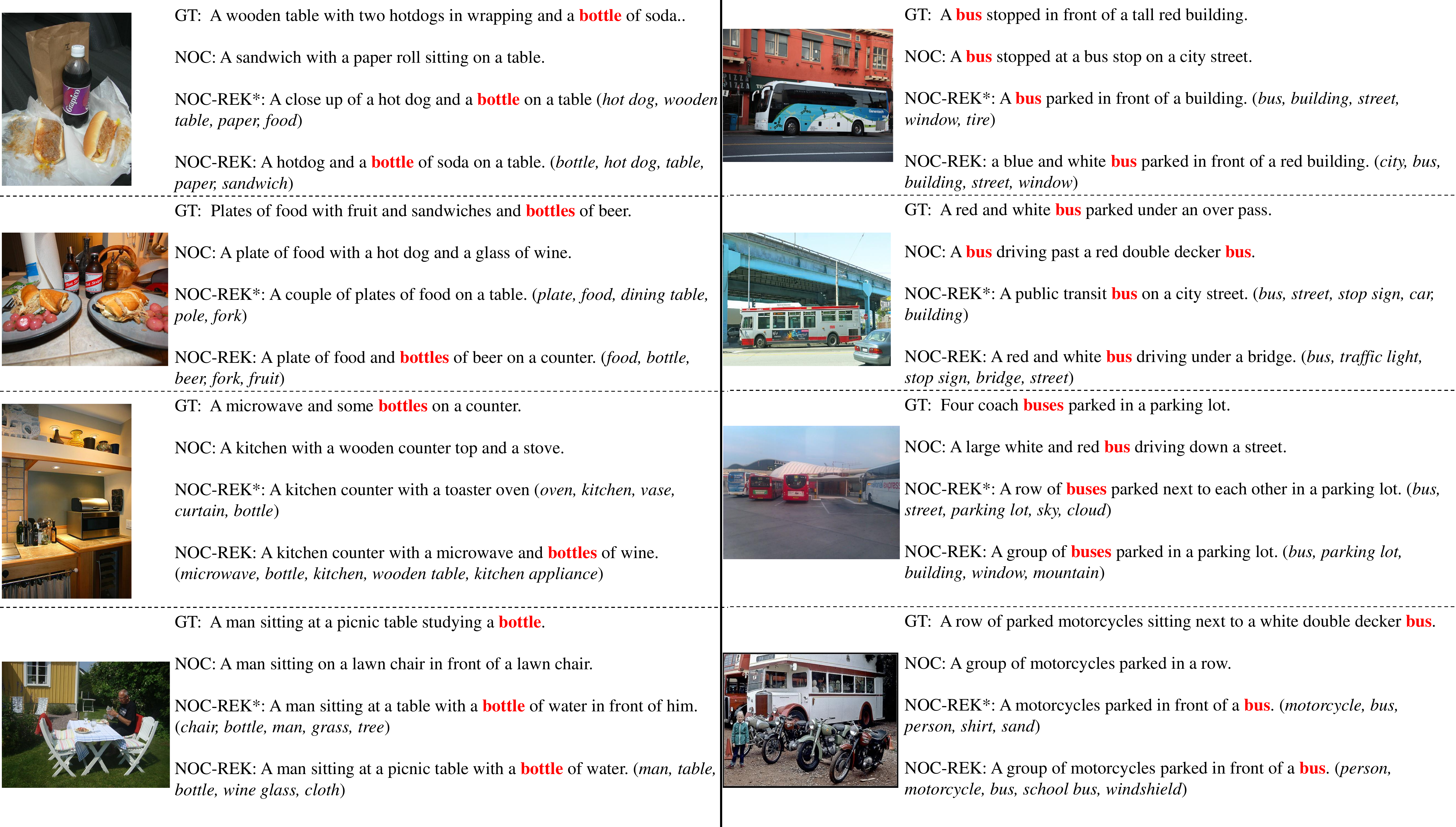}
	\caption{Examples of generated captions by compared methods on held-out COCO, specifically, \red{bottle} and \red{bus} classes. We show the ground-truth captions (GT) for reference. 
	NOC~\cite{venugopalan17noc} usually fails to generate captions with novel objects.
	Our \ours{}, on the other hand, successfully generates correct, fluent, and coherent captions with novel objects. 
	Words in parentheses are top-5 retrieved vocabulary by our method that are reasonably related to objects in image. \red{Red} texts indicate novel objects in the captions.} 
	\label{fig:heldout_1}
\end{figure*}

\begin{figure*}[tb]
	\centering
	\includegraphics[width=1\linewidth]{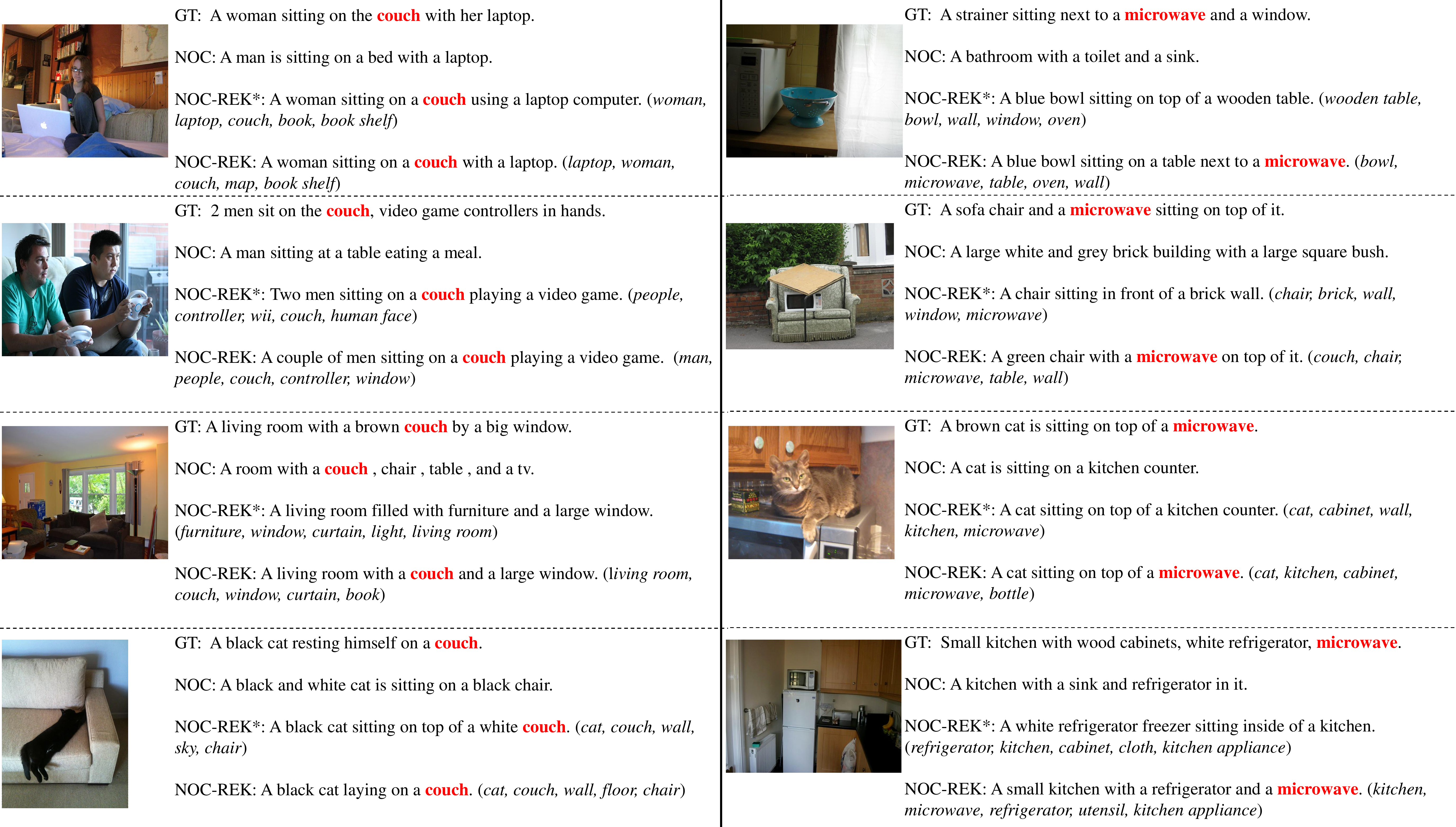}
	\caption{Examples of generated captions by compared methods on held-out COCO, specifically, \red{couch} and \red{microwave} classes. We show the ground-truth captions (GT) on for reference. 
	NOC~\cite{venugopalan17noc} usually fails to generate captions with novel objects.
	Our \ours{}, on the other hand, successfully generates correct, fluent, and coherent captions with novel objects. 
	Words in parentheses are top-5 retrieved vocabulary by our method that are reasonably related to objects in image. \red{Red} texts indicate novel objects in the captions.} 
	\label{fig:heldout_2}
\end{figure*}

\begin{figure*}[tb]
	\centering
	\includegraphics[width=1\linewidth]{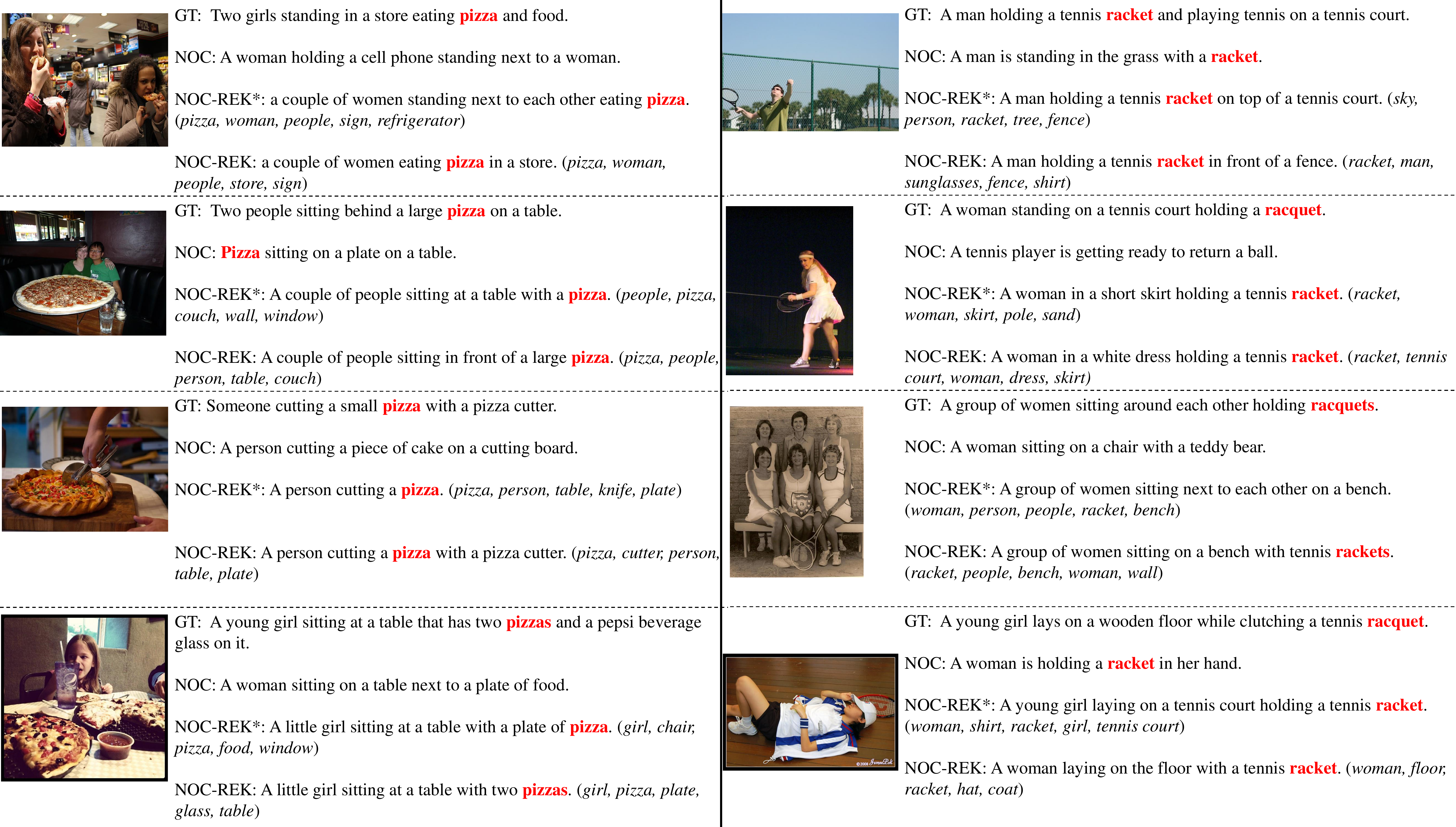}
	\caption{Examples of generated captions by compared methods on held-out COCO, specifically, \red{pizza} and \red{racket} classes. We show the ground-truth captions (GT) for reference. 
	NOC~\cite{venugopalan17noc} usually fails to generate captions with novel objects.
	Our \ours{}, on the other hand, successfully generates correct, fluent, and coherent captions with novel objects. 
	Words in parentheses are top-5 retrieved vocabulary by our method that are reasonably related to objects in image. \red{Red} texts indicate novel objects in the captions.} 
	\label{fig:heldout_3}
\end{figure*}

\begin{figure*}[tb]
	\centering
	\includegraphics[width=1\linewidth]{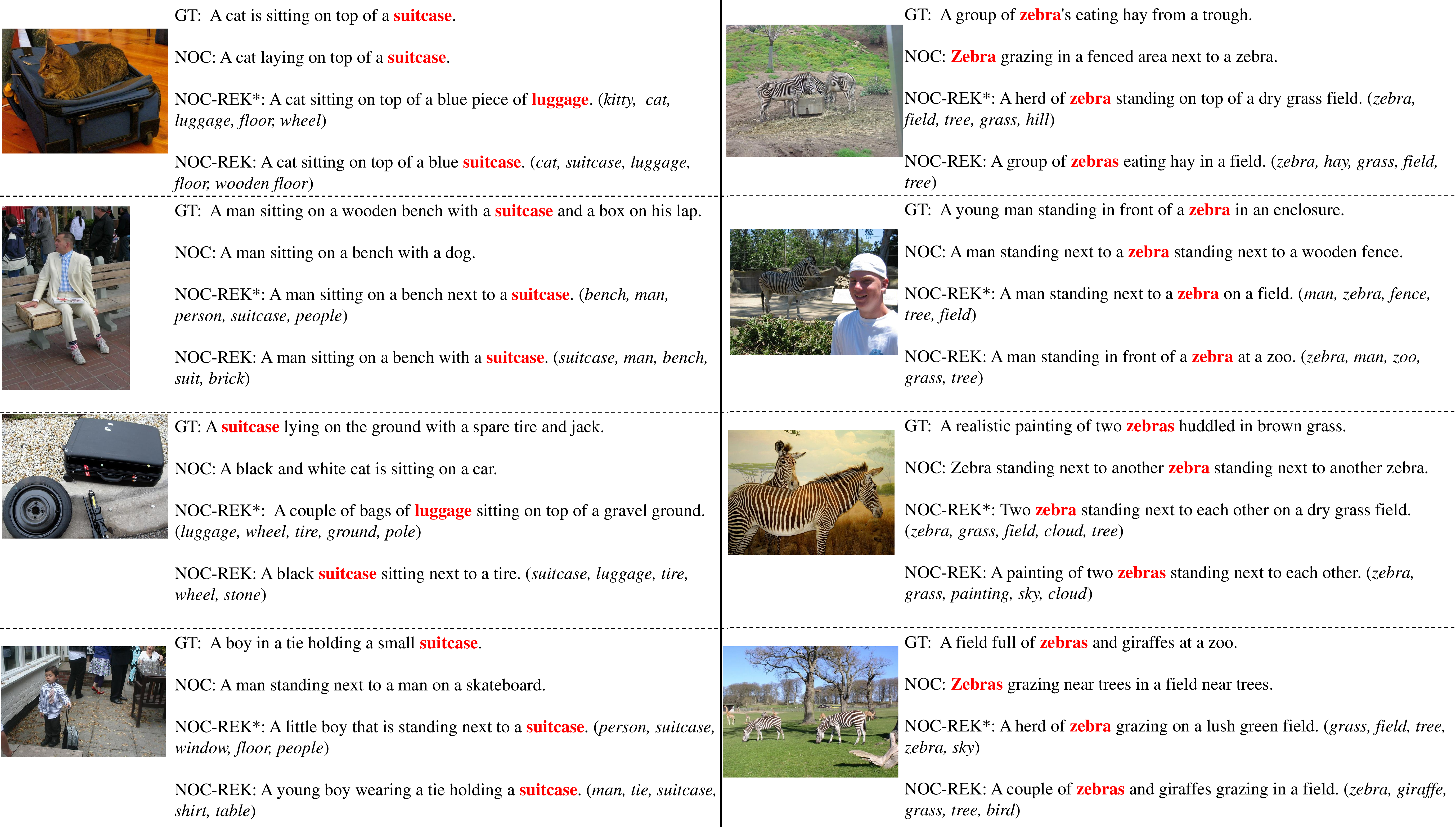}
	\caption{Examples of generated captions by compared methods on held-out COCO, specifically, \red{suitcase} and \red{zebra} classes. We show the ground-truth captions (GT) for reference. 
	NOC~\cite{venugopalan17noc} usually fails to generate captions with novel objects.
	Our \ours{}, on the other hand, successfully generates correct, fluent, and coherent captions with novel objects. 
	Words in parentheses are top-5 retrieved vocabulary by our method that are reasonably related to objects in image. \red{Red} texts indicate novel objects in the captions.} 
	\label{fig:heldout_4}
\end{figure*}

We show more examples of generated captions on held-out COCO dataset~\cite{hendricks16dcc} in Figs.~\ref{fig:heldout_1},~\ref{fig:heldout_2},~\ref{fig:heldout_3}, and~\ref{fig:heldout_4}.
We can see that our method successfully retrieves novel objects and includes them in the captions in a sensible fashion. 
Meanwhile, NOC~\cite{venugopalan17noc} usually fails to generate a caption with a novel object or generates pretty weird captions.
Furthermore, in Fig.~\ref{fig:heldout_2} (right), the ground truth captions sometimes include \textit{racquet}, explaining the mismatching between our method and ground truth dataset as we mentioned in main manuscript.

\begin{figure*}[tb]
	\centering
	\includegraphics[width=1\linewidth]{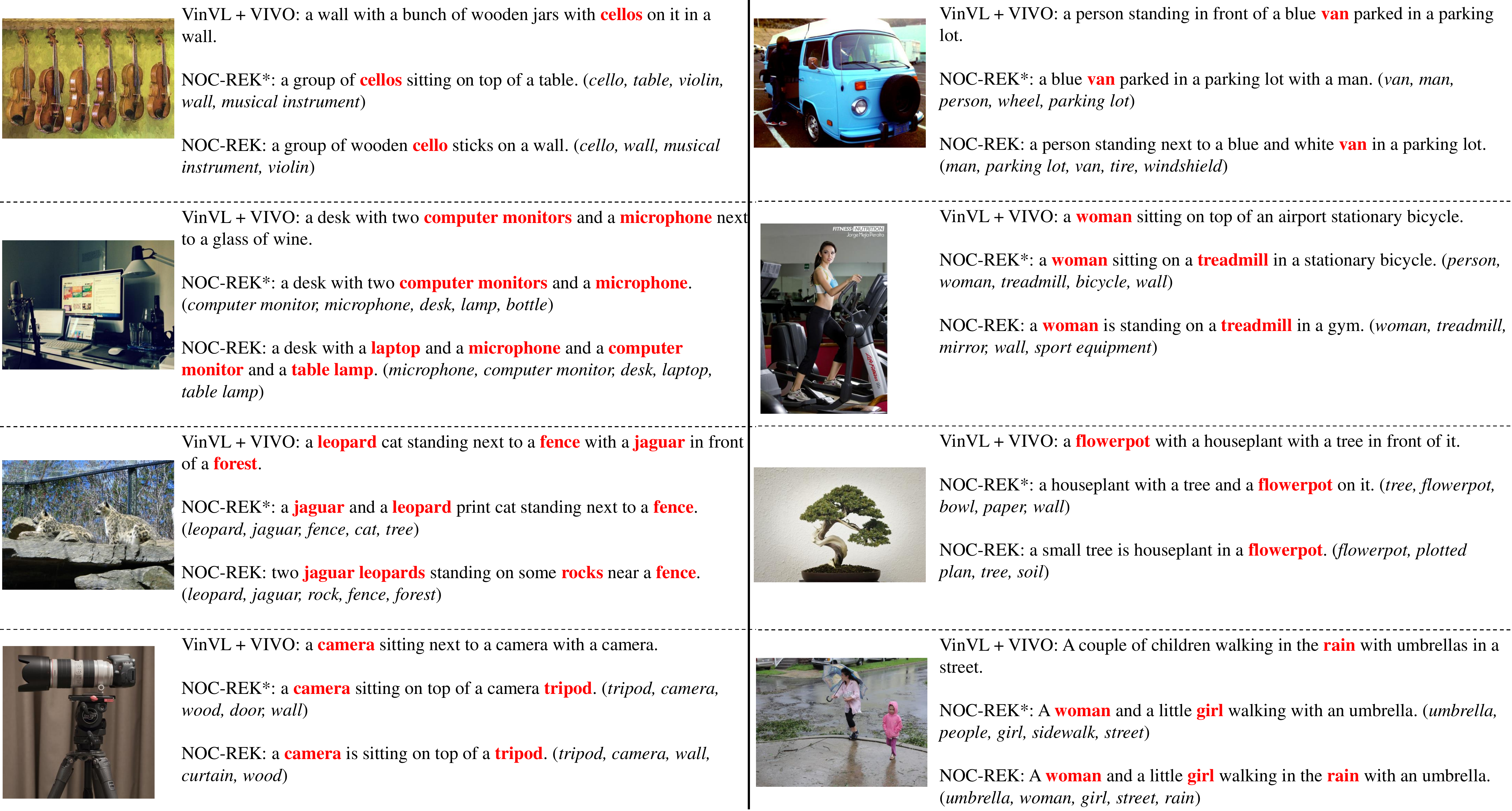}
	\caption{Examples of generated captions by compared methods Nocaps. 
	VinVL+VIVO~\cite{zhang2021vinvl,hu2021vivo} sometimes cannot include the novel objects in the captions.
	Our \ours{}, on the other hand, successfully generates correct, fluent, and coherent captions with novel objects. 
	Words in parentheses are top-5 retrieved vocabulary by our method that are reasonably related to objects in image. \red{Red} texts indicate novel objects in the captions.} 
	\label{fig:nocaps_1}
\end{figure*}

\begin{figure*}[tb]
	\centering
	\includegraphics[width=1\linewidth]{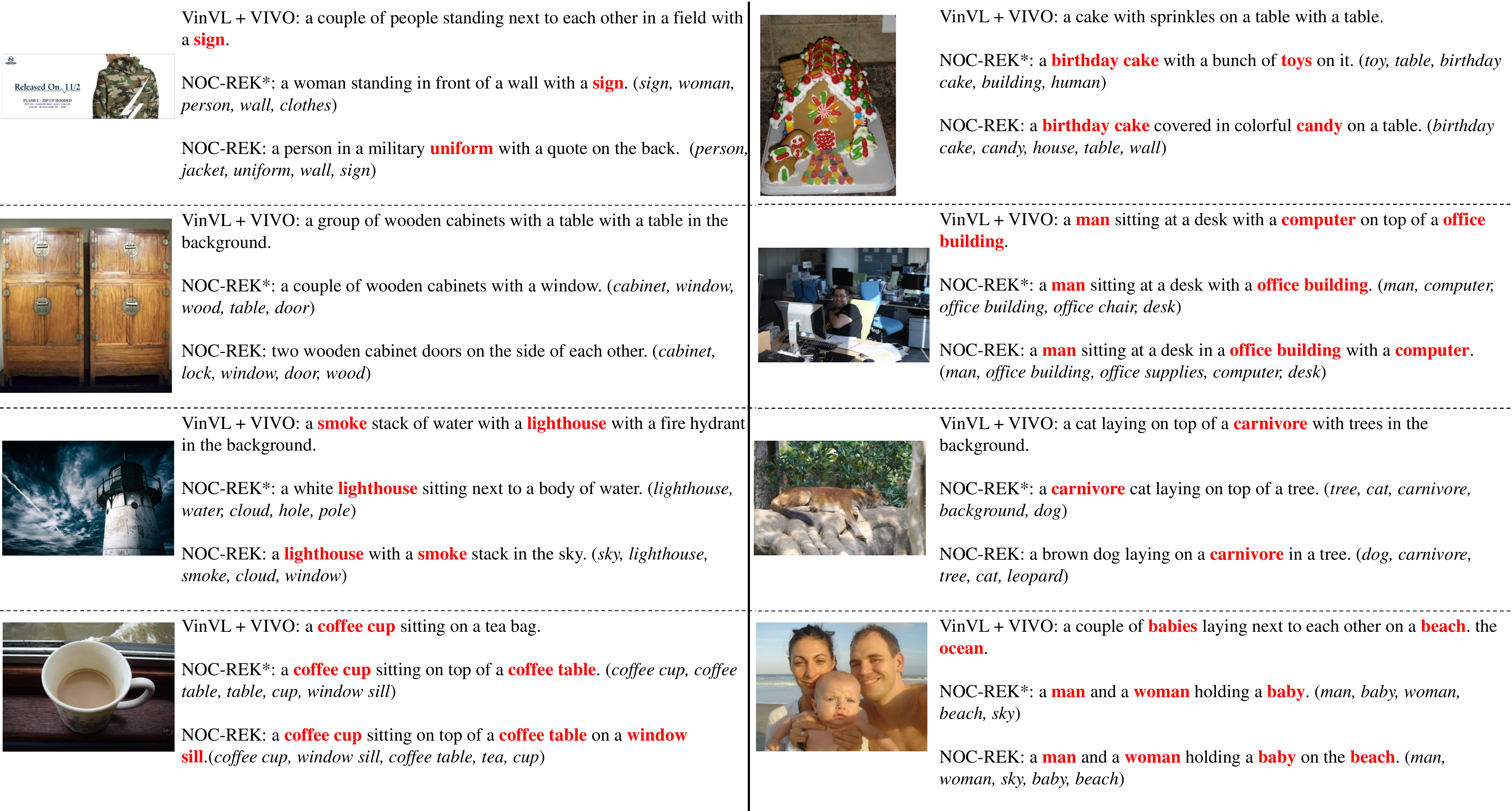}
	\caption{Examples of generated captions by compared methods Nocaps. 
	VinVL+VIVO~\cite{zhang2021vinvl,hu2021vivo} sometimes cannot include the novel objects in the captions.
	Our \ours{}, on the other hand, successfully generates correct, fluent, and coherent captions with novel objects. 
	Words in parentheses are top-5 retrieved vocabulary by our method that are reasonably related to objects in image. \red{Red} texts indicate novel objects in the captions.} 
	\label{fig:nocaps_2}
\end{figure*}

We further show generated captions obtained by \ours{} and VinVL+VIVO~\cite{zhang2021vinvl,hu2021vivo} on Nocaps dataset~\cite{Agrawal2019nocaps} in Figs.~\ref{fig:nocaps_1} and~\ref{fig:nocaps_2}.
Occasionally, VinVL+VIVO~\cite{zhang2021vinvl,hu2021vivo} cannot generate captions consisting of novel objects or a complete sentence.
Generally, our method is capable of generating captions with more objects than those by VinVL+VIVO~\cite{zhang2021vinvl,hu2021vivo}.

\end{document}